\documentclass{article}
\usepackage{spconf,amsmath,epsfig}
\usepackage[noadjust]{cite}
\usepackage{amssymb}
\usepackage{amsfonts}
\usepackage{mathtools}
\usepackage{bbm}
\usepackage{pifont}
\def\cmark{\ding{51}} 
\def\xmark{\ding{55}}
\usepackage{upgreek}
\usepackage{float}
\usepackage{subcaption}
\usepackage{multirow}
\usepackage{verbatim}

\DeclareMathAlphabet\mathbfcal{OMS}{cmsy}{b}{n}
\title{Learning Across Decentralized Multi-Modal Remote Sensing Archives with Federated Learning}
\name{Bar{ı}\c{s} B\"{u}y\"{u}kta\c{s}{\normalfont\textsuperscript{1,2}}, Gencer Sumbul{\normalfont\textsuperscript{1}} and Beg\"{u}m Demir{\normalfont\textsuperscript{1,2}}}
\address{\textsuperscript{1}Faculty of Electrical Engineering and Computer Science, Technische Universit\"at Berlin, Germany\\
\textsuperscript{2}BIFOLD - Berlin Institute for the Foundations of Learning and Data, Germany}
\begin{document}
%
\maketitle

\begin{abstract}
The development of federated learning (FL) methods, which aim to learn from distributed databases (i.e., clients) without accessing data on clients, has recently attracted great attention. Most of these methods assume that the clients are associated with the same data modality. However, remote sensing (RS) images in different clients can be associated with different data modalities that can improve the classification performance when jointly used. To address this problem, in this paper we introduce a novel multi-modal FL framework that aims to learn from decentralized multi-modal RS image archives for RS image classification problems. The proposed framework is made up of three modules: 1) multi-modal fusion (MF); 2) feature whitening (FW); and 3) mutual information maximization (MIM). The MF module performs iterative model averaging to learn without accessing data on clients in the case that clients are associated with different data modalities. The FW module aligns the representations learned among the different clients. The MIM module maximizes the similarity of images from different modalities. Experimental results show the effectiveness of the proposed framework compared to iterative model averaging, which is a widely used algorithm in FL. The code of the proposed framework is publicly available at https://git.tu-berlin.de/rsim/MM-FL.


\end{abstract}
\section{Introduction}
\label{sec:intro}

\begin{figure}[t]
\centering
\centerline{\includegraphics[width=0.85\columnwidth,keepaspectratio]{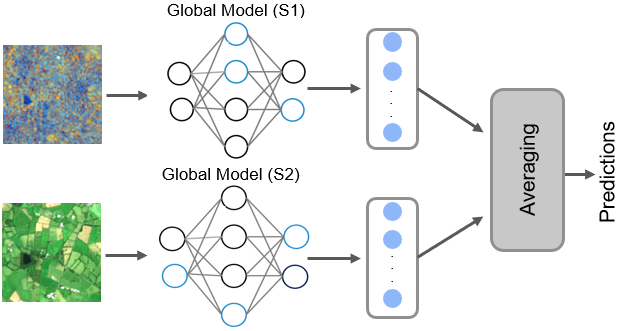}}
\caption{An illustration of MSFedAvg during inference, while the clients are assumed to separately include Sentinel-1 (S1) and Sentinel-2 (S2) images for the sake of simplicity.}
\label{figure}
\end{figure}

Remote sensing (RS) image archives can be stored under different databases due to their growth in size and the data storage limitations of gathering all the data in a centralized server. In addition, some RS archives of data providers (e.g., commercial providers) may not be directly accessible due to commercial concerns, legal regulations, etc. Legal restrictions, such as privacy laws and national security concerns, may also prohibit public access to sensitive information present in the RS image archives \cite{zhang2022progress}. However, most of the deep learning (DL) based approaches require full access to data, while learning the model parameters of deep neural networks (DNNs) during training. To overcome these challenges, federated learning (FL) can be used when there is no access to data on decentralized RS image archives. FL aims to learn DNN models on distributed databases (i.e., clients) and to find the optimal model parameters in a global server (i.e., global model) without accessing data on clients. As one of the first FL studies, the federated averaging (FedAvg) algorithm is introduced in~\cite{FedAvg} to learn a global model by iterative model averaging. In this algorithm, a local model is trained on each client and then its parameters are sent to the global server, in which the parameters of all local models are iteratively averaged and sent back to clients. Although FL is highly studied in computer vision \cite{FedAvg, li2021model,ma2022layer,gao2022feddc}, it is seldom considered in RS \cite{zhang2022prototype}. As an example, in \cite{zhang2022prototype} an FL algorithm is proposed to learn several global models through hierarchical clustering (denoted as FedPHC) when RS images are non-independently and identically distributed among clients. FedPHC assumes that RS images in the clients are associated with the same data modality. However, RS images on different clients can be associated with different data modalities. In addition, multi-modal images associated with the same geographical area allow for a rich characterization of RS images when jointly considered, and thus improve the effectiveness of the considered image analysis task. To jointly exploit multi-modal RS images, the development of image classification methods has attracted great attention in RS \cite{hong2020more,wu2021convolutional,sumbul2022novel}. However, these methods assume that all the multi-modal RS images are accessible during training and thus cannot be directly utilized in the framework of FL. In addition, the adaptation of well-known FL algorithms for the cases, where RS images in different clients are associated with different data modalities, may not be always feasible. To adapt iterative model averaging for such cases, one could operate a dedicated FedAvg algorithm for each modality during training and then employ late fusion during inference (denoted as MSFedAvg). In this way, there is one global model learnt for each modality. Fig.~\ref{figure} shows the inference phase of MSFedAvg. As one can see from the figure, the images on the same geographical area are fed to the global models based on their modality. Then, the prediction is made by averaging the resulting class probabilities. However, MSFedAvg does not extract and exploit the complementary information among the different modalities during the training stage. This may result in limited image classification performance. We would like to note that most of the  existing FL algorithms assume that the same DNN architecture is considered in each client. However, different DNN architectures can be required for RS images with different data modalities. 

To address these issues, in this paper we propose a novel multi-modal FL framework that aims to accurately learn model parameters from decentralized multi-modal RS image archives without having direct access to images in the clients for RS image classification problems.
\section{Proposed Multi-Modal Federated Learning Framework for RS\\Image Classification}

Let $K$ be the total number of clients and $C^i$ be the $i$th client, where $1\leq i\leq K$. Each client $C^i$ locally holds the corresponding training set $D_i = \{(\boldsymbol{x}_{z}^i, \boldsymbol{y}_{z}^i)\}_{z=1}^{M^i}$ including $M^i$ samples, where $\boldsymbol{x}_{z}^i$ is the $z$th RS image of the $i$th client, and $\boldsymbol{y}_{z}^i$ is the corresponding class label. Let $\phi^i$ and $\mu^i$ be the image encoder and classifier, respectively, which are employed for local training on $D^i$. We assume that data modalities can be different between clients (i.e., data modalities associated with $\boldsymbol{x}_{z}^i$ and $\boldsymbol{x}_{z}^j$ can be different for $i\neq j \!\!\!\!\quad \forall z$). The proposed framework aims to learn DNNs on a central server for RS image classification problems without accessing data on clients, while different clients can contain RS images associated with different data modalities. To this end, our framework includes three modules: 1) multi-modal fusion (MF); 2) feature whitening (FW); and 3) mutual information maximization (MIM). Fig.~2 shows an illustration of the local training stage of the proposed framework, which is explained in detail in the following.

\begin{figure}[t]
\centering
\begin{subfigure}[b]{0.48\textwidth}
   \includegraphics[width=1\columnwidth,keepaspectratio]{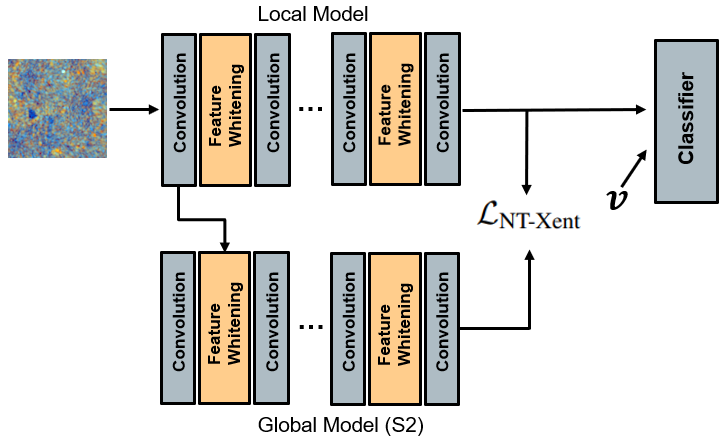}
   \caption{}
   \label{fig:Ng1} 
\end{subfigure}
\begin{subfigure}[b]{0.48\textwidth}
   \includegraphics[width=1\columnwidth,keepaspectratio]{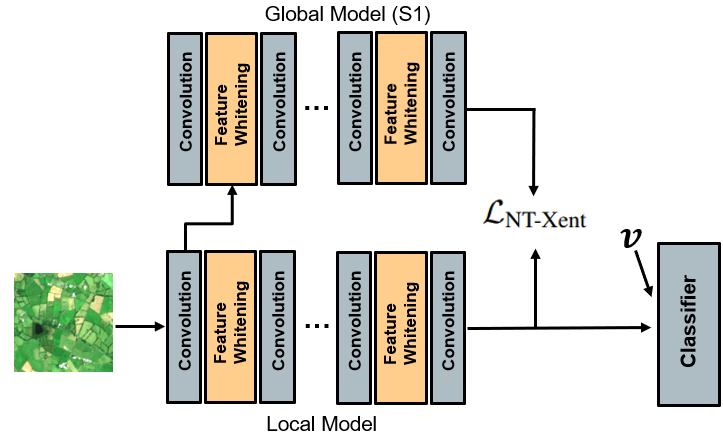}
   \caption{}
   \label{fig:Ng2}
\end{subfigure}
\caption{An illustration of the local training stage of our framework when two clients (a) and (b) are considered and include images associated with Sentinel-1 (S1) and Sentinel-2 (S2), respectively, for the sake of simplicity.}
\end{figure}
The MF module aims to employ DNNs for iterative model averaging when clients are associated with different data modalities. To this end, instead of employing a single DNN, we define modality-specific backbones and a common classifier on the server side. Let $P$ be the total number of modalities associated with clients. Let $\psi_m$ be a modality-specific backbone while $1\leq m\leq P \leq K$ and $\theta$ be a common classifier. After defining DNNs on the server side, the model parameters are shared with the clients. Then, $\mu^i$ is updated with $\theta$ for the $i$th client and $\phi^i$ is updated with the corresponding modality-specific backbone. To jointly utilize information from different modalities during inference, the feature vectors extracted by all modality-specific backbones $\{\psi_m\}_{m=1}^P$ are fused through concatenation. However, during local training, a pseudo fusion via concatenation with the zero vector $\boldsymbol{v}$ = $\vec{0}$ is applied since there is no access to data from different clients. Then, the resulting vector is fed into $\mu^i$. MF module allows defining different architectures for different modalities and also jointly utilizing and extracting information from multiple modalities. 

It is worth noting that the distributions of feature vectors may differ between clients because of the modality differences that may lead to sub-optimal parameters of the aggregated models. To reduce such distribution differences, the FW module aims to project the feature vectors of the images with different modalities that are extracted by local models into a common distribution. This is achieved by using batch whitening layers instead of batch normalization layers inspired by \cite{roy2019unsupervised}. The batch whitening layer $BW$ for $\boldsymbol{x}_{z,p}^i$ is defined as follows:

\begin{equation}
\begin{split}
\boldsymbol{\hat{x}}^i_{z} &= W_B (\boldsymbol{x}^i_{z} - \mu_B),\\
BW ({\boldsymbol{x}}_{z,p}^i) &= \gamma_p \boldsymbol{\hat{x}}^i_{z,p} + \beta_p,
\end{split}
\label{fw}
\end{equation}
where ${\boldsymbol{x}}_{z,p}^i$ is the $p$th element of ${\boldsymbol{x}}_{z}^i$, $\mu_B$ is the mean vector, $\gamma_p$ is the scaling parameter and $\beta_p$ is the shifting parameter. This module aligns the data distributions between clients using covariance matrices. It uses domain-specific alignment layers, which compute domain-specific covariance matrices of intermediate features. 

The MIM module aims to model mutual-information between different modalities by maximizing the similarity of images, which are acquired on the same geographical area and associated to different modalities. To achieve this, we employ the NT-Xent loss function $\mathcal{L}_{NTX}$~\cite{Chen:2020} for modelling similarity through the feature vectors of a local model and that of aggregated models from different modalities. To this end, the feature vectors are extracted by feeding images into the local and global models. Since the global models are not trained with images that have the same modality as the images in local client, it is not possible to directly feed into the global model. Therefore, images are fed to the first convolutional layer of the local model and the output is fed to the global models. Then, the $\mathcal{L}_{NTX}$ for a given mini-batch $\mathbfcal{B}$ is calculated using two feature vectors as follows:
\begin{equation}
\mathcal{L}_{NTX}(\mathbfcal{B}) \! = \!\! - \!\!\! \sum_{\boldsymbol{x}_z^i \in \mathbfcal{B}} \!\!  \log \frac{ e^{S(  \phi^i(\boldsymbol{x}_z^i) ,  \psi_m(\boldsymbol{x}_z^i)  ) / \tau } } {\sum\limits_{\boldsymbol{x}_t^i \in \mathbfcal{B}} \mathbbm{1}_{[z \neq t]} e^{S(  \phi^i(\boldsymbol{x}_z^i) ,  \psi_m(\boldsymbol{x}_t^i)  ) / \tau } },
\end{equation}
where $\tau$ is a temperature parameter, $\mathbbm{1}$ is the indicator function, $S(.,.)$ measures cosine similarity and $\mathbin\Vert$ is the concatenation operator. Accordingly, we define the local objective $\mathcal{L}$ based on $\mathcal{L}_{NTX}$ and cross-entropy loss $\mathcal{L}_{BCE}$ as follows:
\begin{equation}
    \mathcal{L}(\mathbfcal{B}) \!=\! \mathcal{L}_{NTX}\! (\mathbfcal{B}) \!+\!\!\!\!\!\!  \sum_{\boldsymbol{x}_z^i, \boldsymbol{y}_z^i \in \mathbfcal{B}} \!\!\!\!\! \mathcal{L}_{BCE}(\theta ( \phi^i(\boldsymbol{x}_z^i) \!\! \mathbin\Vert \! \boldsymbol{v}),\boldsymbol{y}_z^i).
\label{moon1}
\end{equation}
Once the local training procedure of our framework is completed, the considered local parameters are sent back to the server and aggregated on the server side.

\section{Experimental Results}

The experiments were conducted on the BigEarthNet-MM archive \cite{sumbul2021bigearthnet}. It includes 590,326 multi-modal image pairs, each of which includes Sentinel-1 and Sentinel-2 images acquired on the same geographical area. Each pair in BigEarthNet-MM is annotated with multi-labels. The multi-modal image pairs acquired in summer were used for experiments. We defined six clients, in which three clients contain Sentinel-1 images and the remaining clients contain Sentinel-2 images. All six clients participate in each round of the local training. We defined two different scenarios: 1) the images are randomly distributed to the clients; and 2) the images acquired in the same country are present in the same client. We utilized the ResNet-50 CNN architecture \cite{he2016deep} as the backbone of the proposed framework. We used the Adam optimizer with the learning rate of 0.001 and the momentum of 0.9. We trained our framework for 40 epochs with the mini-bath size of 256. We evaluate the performance of our method in terms of classification accuracy (in $F\textsubscript{1}$-Score) and local training complexity (in seconds).

\begin{table}[t]
\caption{$F_1$-scores (\%) associated with the different combinations of the modules of the proposed framework.}
\centering
\label{tab:1} 
\begin{tabular}{|ccc|cc|}
\hline
\multicolumn{3}{|c|}{\textbf{Module} }                           & \multicolumn{2}{c|}{\textbf{$\boldsymbol{F_1}$-Score}}           \\ \hline
\multicolumn{1}{|c|}{MF} & \multicolumn{1}{c|}{FW} & MIM & \multicolumn{1}{c|}{Scenario 1} & Scenario 2 \\ \hline
\multicolumn{1}{|c|}{\cmark}   & \multicolumn{1}{c|}{\xmark}   &  \xmark   & \multicolumn{1}{c|}{74.4}       & 61.7       \\ \hline
\multicolumn{1}{|c|}{\cmark}   & \multicolumn{1}{c|}{\xmark}   &  \cmark   & \multicolumn{1}{c|}{74.2}       & 67.1       \\ \hline
\multicolumn{1}{|c|}{\cmark}   & \multicolumn{1}{c|}{\cmark}   &  \xmark   & \multicolumn{1}{c|}{75.5}       & 68.5       \\ \hline
\multicolumn{1}{|c|}{\cmark}   & \multicolumn{1}{c|}{\cmark}   &  \cmark   & \multicolumn{1}{c|}{76.7}       & 69.3       \\ \hline
\end{tabular}
\end{table}

In the first set of trials, we analyze the effectiveness of each module of our framework. Table \ref{tab:1} shows the corresponding results. One can see from the table that jointly using three modules achieves the highest results than other combinations for both scenarios. 
This shows that FW and MIM modules reduce the data distribution differences between clients, which leads to accurate characterization of multi-modal RS images during training. The lowest accuracy is obtained by the joint use of MF and MIM modules for Scenario 1, which is 2.5\% lower than that obtained by the joint use of all modules. Moreover, the lowest accuracy is obtained by using only MF module for Scenario 2, which is 7.6\% lower than the joint use of all modules. This indicates that the FW module increases the performance more than the MIM module. One can also observe from the table that the results achieved with Scenario 1 are higher than those achieved with Scenario 2. As an example, using only the MF module provides 74.4\% and 61.7\% accuracies for Scenario 1 and Scenario 2, respectively. This is due to the higher non-IID level of Scenario 2 compared Scenario 1.

\begin{table}[t]
\caption{$F_1$-scores (\%) and local training complexity (in seconds) obtained by MSFedAvg and the proposed framework.}
\centering
\resizebox{\columnwidth}{!}{%
\label{tab:2}
\begin{tabular}{|l|cc|c|}
\hline
\multirow{2}{*}{\textbf{Method} }                                  & \multicolumn{2}{c|}{\textbf{$\boldsymbol{F_1}$-Score}}                & \multirow{2}{*}{\textbf{\begin{tabular}[c]{@{}c@{}}Local Training\\ Complexity\end{tabular}}} \\ \cline{2-3}
                                                           & \multicolumn{1}{c|}{Scenario 1} & Scenario 2 &                                                                                      \\ \hline
MSFedAvg~\cite{FedAvg}                                                   & \multicolumn{1}{c|}{67.2}       & 58.9       &   10.5                                                                                   \\ \hline
\begin{tabular}[c]{@{}l@{}}Proposed \\ Framework\end{tabular} & \multicolumn{1}{c|}{76.7}       & 69.3       &   11.6                                                                                   \\ \hline
\end{tabular}
}
\end{table}

In the second set of trials, we compare the proposed method with MSFedAvg in terms of accuracy and local training complexity under both scenarios. The local training complexity refers to the average computational cost for training the considered DNN for each client. Table \ref{tab:2} shows the corresponding results. One can observe from the table that our framework achieves the highest $F\textsubscript{1}$-scores under both scenarios compared to MSFedAvg. The proposed framework outperforms MSFedAvg by 9.5\% and 10.4\% for Scenario 1 and Scenario 2, respectively. This shows the effectiveness of our framework compared to MSFedAvg. One can also see from the table that the proposed framework leads to higher accuracies than MSFedAvg at the cost of a slight increase in local training complexity. The average completion time of a training round on one client increases only 10\% when the proposed framework is used instead of MSFedAvg. This increase can be compensated by using higher mini-batch sizes to reduce the local training complexity of our framework.

\section{Conclusion}

In this paper, we have introduced a novel framework, which is capable of learning DNN model parameters from decentralized multi-modal RS image archives without accessing data on clients in the context of RS image classification. Our framework includes: i) multi-modal fusion module to perform iterative model averaging when the images in different clients are associated to different data modalities; ii) feature whitening module to align the representations learned among different clients; and iii) mutual information maximization module to maximize the similarity of images from different modalities. Experimental results show the success of the proposed framework compared to MSFedAvg~\cite{FedAvg}. 

We would like to note that the proposed framework is independent from the number of clients and also the modality-specific backbones being selected for the considered modalities. Thus, any DNN architecture specifically designed for each modality (e.g., attention-based CNNs for very high resolution aerial images) can be utilized in our framework. This can allow to accurately describe modality-specific information content, and thus lead to higher RS image classification performance. As a future work, we plan to investigate the joint use of RS images with socio-economic data (e.g., demographic data, household surveys, etc.) in the framework of FL in RS.

\section{ACKNOWLEDGEMENTS}
This work is supported by the European Research Council (ERC) through the ERC-2017-STG BigEarth Project under Grant 759764 and by the German Research Foundation through the IDEAL-VGI project under Grant 424966858.

\bibliographystyle{IEEEtran}
{\small
\bibliography{strings,refs}}

\begin{thebibliography}{10}
\providecommand{\url}[1]{#1}
\csname url@samestyle\endcsname
\providecommand{\newblock}{\relax}
\providecommand{\bibinfo}[2]{#2}
\providecommand{\BIBentrySTDinterwordspacing}{\spaceskip=0pt\relax}
\providecommand{\BIBentryALTinterwordstretchfactor}{4}
\providecommand{\BIBentryALTinterwordspacing}{\spaceskip=\fontdimen2\font plus
\BIBentryALTinterwordstretchfactor\fontdimen3\font minus
  \fontdimen4\font\relax}
\providecommand{\BIBforeignlanguage}[2]{{%
\expandafter\ifx\csname l@#1\endcsname\relax
\typeout{** WARNING: IEEEtran.bst: No hyphenation pattern has been}%
\typeout{** loaded for the language `#1'. Using the pattern for}%
\typeout{** the default language instead.}%
\else
\language=\csname l@#1\endcsname
\fi
#2}}
\providecommand{\BIBdecl}{\relax}
\BIBdecl

\bibitem{zhang2022progress}
B.~Zhang, Y.~Wu, B.~Zhao, J.~Chanussot, D.~Hong, J.~Yao, and L.~Gao, ``Progress
  and challenges in intelligent remote sensing satellite systems,'' \emph{IEEE
  Journal of Selected Topics in Applied Earth Observations and Remote Sensing},
  2022.

\bibitem{FedAvg}
H.~B. McMahan, E.~Moore, D.~Ramage, S.~Hampson, and B.~A. Arcas,
  ``Communication-efficient learning of deep networks from decentralized
  data,'' \emph{International Conference on Artificial Intelligence and
  Statistics}, pp. 1273--1282, 2017.

\bibitem{li2021model}
Q.~Li, B.~He, and D.~Song, ``Model-contrastive federated learning,'' \emph{IEEE
  Conference on Computer Vision and Pattern Recognition}, pp. 10\,713--10\,722,
  2021.

\bibitem{ma2022layer}
X.~Ma, J.~Zhang, S.~Guo, and W.~Xu, ``Layer-wised model aggregation for
  personalized federated learning,'' \emph{IEEE Conference on Computer Vision
  and Pattern Recognition}, pp. 10\,092--10\,101, 2022.

\bibitem{gao2022feddc}
L.~Gao, H.~Fu, L.~Li, Y.~Chen, M.~Xu, and C.~Xu, ``{FedDC}: Federated learning
  with non-iid data via local drift decoupling and correction,'' \emph{IEEE
  Conference on Computer Vision and Pattern Recognition}, pp. 10\,112--10\,121,
  2022.

\bibitem{zhang2022prototype}
B.~Zhang, X.~Zhang, M.~Pun, and M.~Liu, ``Prototype-based clustered federated
  learning for semantic segmentation of aerial images,'' \emph{IEEE
  International Geoscience and Remote Sensing Symposium}, pp. 2227--2230, 2022.

\bibitem{hong2020more}
D.~Hong, L.~Gao, N.~Yokoya, J.~Yao, J.~Chanussot, Q.~Du, and B.~Zhang, ``More
  diverse means better: Multimodal deep learning meets remote-sensing imagery
  classification,'' \emph{IEEE Transactions on Geoscience and Remote Sensing},
  vol.~59, no.~5, pp. 4340--4354, 2020.

\bibitem{wu2021convolutional}
X.~Wu, D.~Hong, and J.~Chanussot, ``Convolutional neural networks for
  multimodal remote sensing data classification,'' \emph{IEEE Transactions on
  Geoscience and Remote Sensing}, vol.~60, pp. 1--10, 2021.

\bibitem{sumbul2022novel}
G.~Sumbul, M.~M{\"u}ller, and B.~Demir, ``A novel self-supervised cross-modal
  image retrieval method in remote sensing,'' \emph{IEEE International
  Conference on Image Processing}, pp. 2426--2430, 2022.

\bibitem{roy2019unsupervised}
S.~Roy, A.~Siarohin, E.~Sangineto, S.~R. Bulo, N.~Sebe, and E.~Ricci,
  ``Unsupervised domain adaptation using feature-whitening and consensus
  loss,'' \emph{IEEE Conference on Computer Vision and Pattern Recognition},
  pp. 9471--9480, 2019.

\bibitem{Chen:2020}
T.~Chen, S.~Kornblith, M.~Norouzi, and G.~Hinton, ``A simple framework for
  contrastive learning of visual representations,'' \emph{International
  Conference on Machine Learning}, pp. 1597--1607, 2020.

\bibitem{sumbul2021bigearthnet}
G.~Sumbul, A.~D.~Wall, T.~Kreuziger, F.~Marcelino, H.~Costa, P.~Benevides,
  M.~Caetano, B.~Demir, and V.~Markl, ``{BigEarthNet-MM}: A large-scale,
  multimodal, multilabel benchmark archive for remote sensing image
  classification and retrieval,'' \emph{IEEE Geoscience and Remote Sensing
  Magazine}, vol.~9, no.~3, pp. 174--180, 2021.

\bibitem{he2016deep}
K.~He, X.~Zhang, S.~Ren, and J.~Sun, ``Deep residual learning for image
  recognition,'' in \emph{IEEE Conference on Computer Vision and Pattern
  Recognition}, 2016, pp. 770--778.

\end{thebibliography}
\end{document}